\documentclass[conference]{IEEEtran}
\IEEEoverridecommandlockouts
\usepackage{cite}
\usepackage{amsmath,amssymb,amsfonts}
\usepackage{graphicx}
\usepackage{textcomp}
\usepackage[table]{xcolor} % 加载 xcolor 宏包
\usepackage{array}
\usepackage[caption=false,font=normalsize,labelfont=sf,textfont=sf]{subfig}
\usepackage{stfloats}
\usepackage{url}
\usepackage{verbatim}
\usepackage{diagbox} % Include the diagbox package
\usepackage{algorithm}
\usepackage{algpseudocode}
\usepackage{hyperref}       % hyperlinks
\usepackage{authblk}

\def\BibTeX{{\rm B\kern-.05em{\sc i\kern-.025em b}\kern-.08em
    T\kern-.1667em\lower.7ex\hbox{E}\kern-.125emX}}
\begin{document}

\title{A Semantic-Enhanced Heterogeneous Graph Learning Method for Flexible Objects Recognition}
% \author{Anonymous ICME submission}

\author[1]{Kunshan Yang}
\author[1]{Wenwei Luo}
\author[1]{Yuguo Hu}
\author[1]{Jiafu Yan}
\author[1]{Mengmeng Jing}
\author[1,*]{Lin Zuo}

\affil[1]{University of Electronic Science and Technology of China

% \author{Kunshan Yang\textsuperscript{1}, Wenwei Luo\textsuperscript{1}, Yuguo Hu\textsuperscript{1}, Jiafu Yan\textsuperscript{2}, Mengmeng Jing\textsuperscript{1}, Lin Zuo\textsuperscript{1,*}

% \thanks{This work was supported by the National
% Natural Science Foundation of China under Grant 62276054 and the
% Sichuan Science and Technology Program under Grant 2023YFG0156.}

\thanks{\textsuperscript{*}Corresponding author: Lin Zuo. Email: linzuo@uestc.edu.cn}
% \thanks{This work was supported by the National
% Natural Science Foundation of China under Grant 62276054 and the National Natural Science Foundation of China under Grant 62406060.}

% \thanks{\textsuperscript{1}School of Information and Software Engineering, University of Electronic Science and Technology of China. \textsuperscript{2}School of Mechanical and Electrical Engineering, University of Electronic Science and Technology of China.}

\thanks{This work was supported by the National Natural Science Foundation of China (62276054) and the National Natural Science Foundation of China (62406060).}
}

% \author{
% \IEEEauthorblockN{%
% Kunshan Yang\IEEEauthorrefmark{1}, 
% Wenwei Luo\IEEEauthorrefmark{1}, 
% Yuguo Hu\IEEEauthorrefmark{1}, 
% Jiafu Yan\IEEEauthorrefmark{2}, 
% Mengmeng Jing\IEEEauthorrefmark{1}, 
% Lin Zuo\IEEEauthorrefmark{1}%
% }
% \IEEEauthorblockA{\IEEEauthorrefmark{1}School of XX, 
% University of Electronic Science and Technology of China, 
% Chengdu 611731, China}
% \IEEEauthorblockA{\IEEEauthorrefmark{2}Department of YY, 
% Some Other University, 
% Some City 100000, China}
% \IEEEauthorblockA{Email: linzuo@uestc.edu.cn}
% }

% 1. Heterogeneous Graph Learning and Adaptive Scanning for Flexible Objects Recognition
% 2. Flexible Object Recognition: Bridging Semantic and Visual Features with   Adaptive Scanning and Heterogeneous Graph Learning
% 3. A Semantic-Enhanced Heterogeneous Graph Learning Method for Flexible Object Recognition
% 4. Semantic and Visual Alignment in Flexible Object Recognition: A Heterogeneous Graph Methodattention mechanism

\maketitle

\begin{abstract}

Flexible objects recognition remains a significant challenge due to its inherently diverse shapes and sizes, translucent attributes, and subtle inter-class differences. Graph-based models, such as graph convolution networks and graph vision models, are promising in flexible objects recognition due to their ability of capturing variable relations within the flexible objects. These methods, however, often focus on global visual relationships or fail to align semantic and visual information. To alleviate these limitations, we propose a semantic-enhanced heterogeneous graph learning method. First, an adaptive scanning module is employed to extract discriminative semantic context, facilitating the matching of flexible objects with varying shapes and sizes while aligning semantic and visual nodes to enhance cross-modal feature correlation. Second, a heterogeneous graph generation module aggregates global visual and local semantic node features, improving the recognition of flexible objects. Additionally, We introduce the FSCW, a large-scale flexible dataset curated from existing sources. We validate our method through extensive experiments on flexible datasets (FDA and FSCW), and challenge benchmarks (CIFAR-100 and ImageNet-Hard), demonstrating competitive performance.

\end{abstract}


\section{Introduction}
In the field of multimedia, object recognition has primarily focused on rigid objects \cite{huang2017densely}, such as cups, bottles, and birds. These rigid objects maintain relatively stable shapes and sizes, and can be accurately recognized through Convolutional Neural Networks (CNNs) \cite{huang2017densely, radosavovic2020designing}, Transformers \cite{yuan2021tokens, dosovitskiy2020image, touvron2021training, ding2022davit, chu2021twins}, and other methods \cite{tolstikhin2021mlpmixer}. However, the recognition of flexible objects remains a significant challenge due to their inherently diverse shapes, sizes, translucent attributes and ambiguous boundaries.

Flexible objects, such as clouds, smoke, and other deformable entities, are crucial for various applications, including early fire detection \cite{jadon2019firenet}, weather forecasting and climate monitoring \cite{zhang2018cloudnet}. Despite their importance, existing object recognition methods struggle to cope with the complexities posed by flexible objects \cite{jadon2019firenet, zhang2018cloudnet}.

\begin{figure}[ht]
  \centering
  \includegraphics[width=\linewidth]{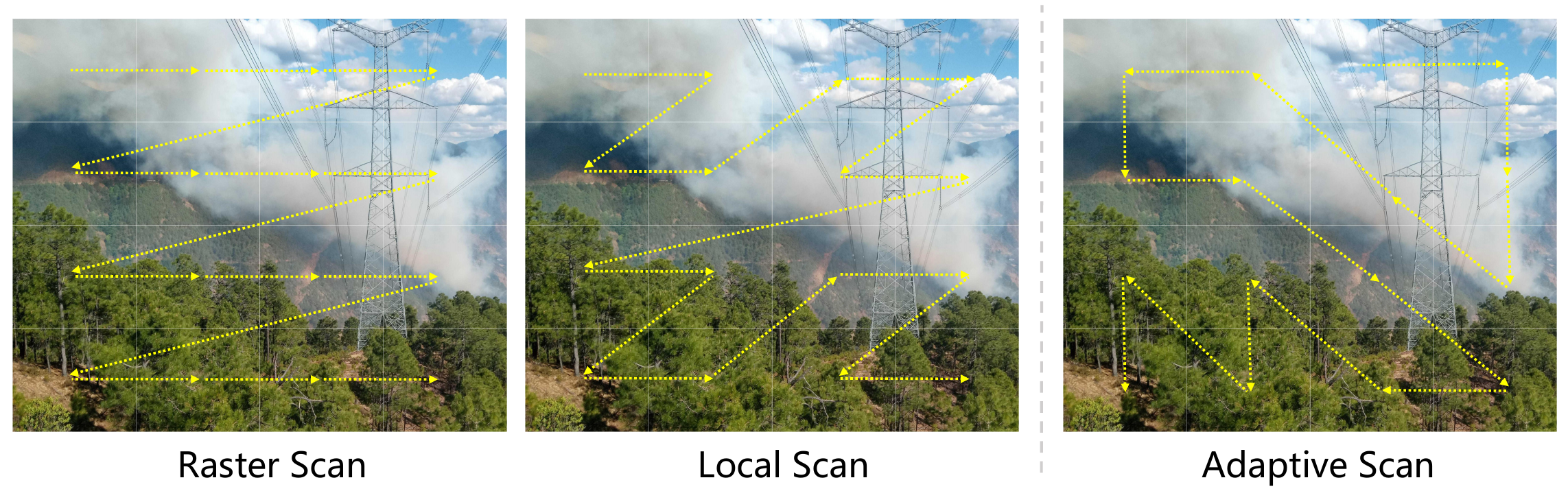}
  \caption{Raster Scan suffer from discontinuities, while Local Scan struggles with semantic irrelevance. In contrast, our adaptive scanning method enables the state space model to better capture semantic node context, extracting more discriminative features for flexible objects of varying shapes and sizes.}
\end{figure}

Graph-based methods, e.g., Graph Convolution Networks (GCNs) and Graph Vision Models (GVM) \cite{yang2025flexible, han2022vision, yao2022ml}, have shown promise in flexible object recognition by modeling the variable relationships between image patches or nodes. However, they often neglect local semantic information and the dynamic correlation between cross-modal features. \cite{yang2025flexible} proposes improved FViG based on Vision GNN \cite{han2022vision}, a patch-level flexible objects recognition method, that adaptively captures relationships between nodes. However, these ViG-based methods \cite{yang2025flexible, han2022vision} only explores the spatial relationship of the global visual area, and ignores the local semantic relationship \cite{li2024deep, yao2022ml}. Additionally, existing graph learning approach \cite{yao2022ml} that consider both global visual and local semantic features has not yet addressed the alignment between semantic and visual nodes, nor the dynamic correlation between cross-modal features.

To address these limitations, we draw inspiration from State-Space Models (SSMs), which have demonstrated effectiveness in extracting local semantic features \cite{huang2024localmamba}. SSMs employ scanning strategies \cite{liu2024vmambavisualstatespace, huang2024localmamba} to map 2D images into 1D sequences, making them suitable for visual tasks. However, these pre-defined scanning strategies are relatively fixed and suffer from issues such as discontinuity in neighboring node space and semantic irrelevance, as illustrated in Fig. 1. Motivated by the strengths of SSMs, we propose an adaptive scanning module into the heterogeneous graph model to overcome these shortcomings. Specifically, the adaptive scanning module enables the effective capture of contextual relationships between semantic nodes, facilitating the extraction of discriminative semantic features to match flexible objects with varying shapes and sizes. This improvement also achieves alignment between semantic and visual nodes, thereby enhancing the dynamic correlation between cross-modal features. Furthermore, we propose a heterogeneous graph generation module, which enables the graph network to aggregate both global visual and local semantic node features. Extensive experiments validate the effectiveness of our method in enhancing the recognition performance of flexible objects. The main contributions are as follows.

\begin{figure*}[ht]
  \centering
  \includegraphics[width=0.90\linewidth]{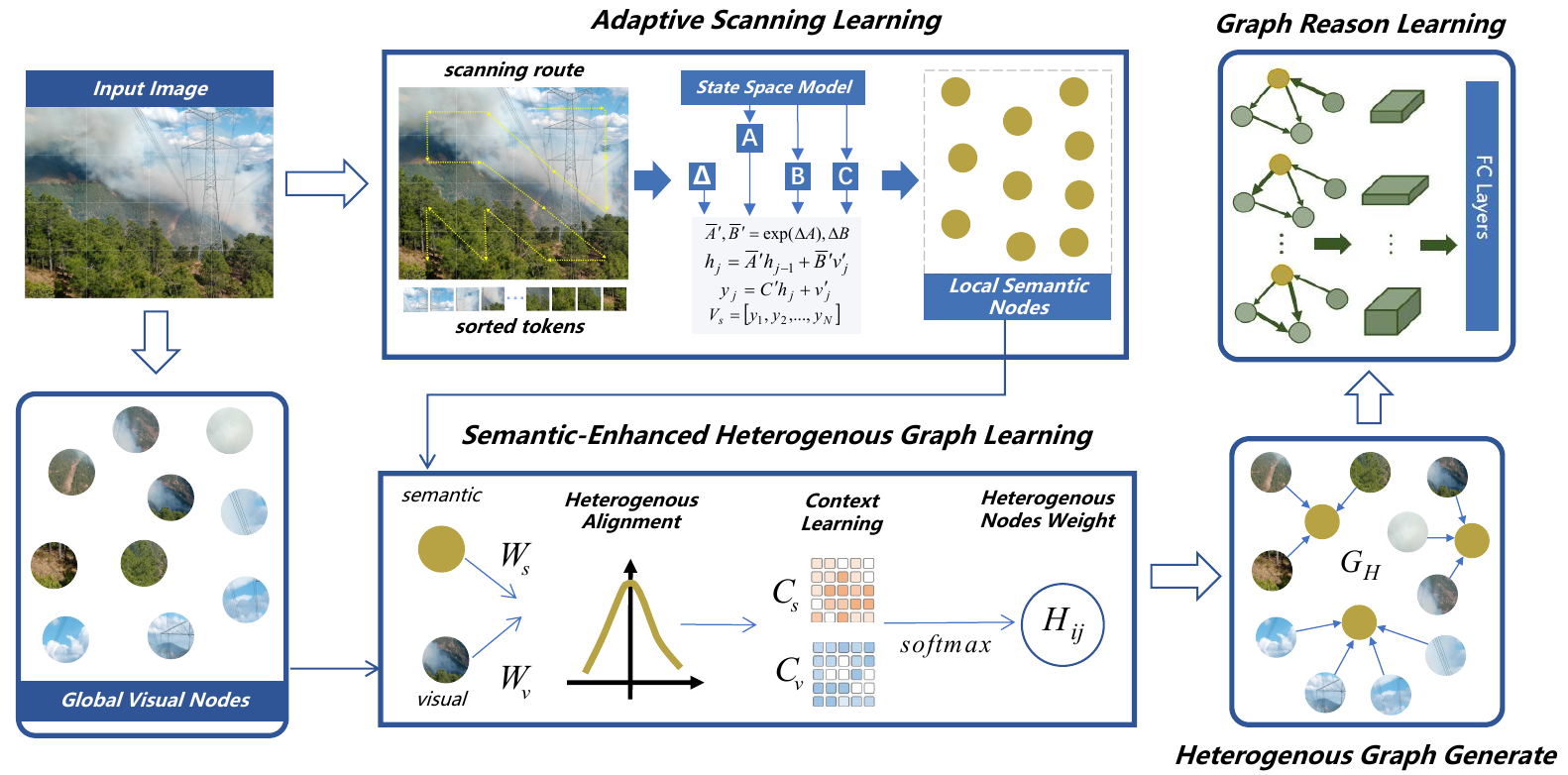}
  \caption{Method overview. Semantic-enhanced heterogeneous graph learning that employs an adaptive scanning module, enables extraction of discriminative semantic context for matching flexible objects with varying shapes and sizes, while aligning semantic and visual nodes to enhance cross-modal feature correlation. Additionally, heterogeneous graph generation allows the graph neural network to effectively aggregate global visual and local semantic node features.}
\end{figure*}

(1) We propose a semantic-enhanced heterogeneous graph learning method that employs an adaptive scanning module into the heterogeneous graph. This facilitates the extraction of discriminative semantic context, enabling the matching of flexible objects with varying shapes and sizes. Furthermore, it achieves alignment between semantic and visual nodes, thereby enhancing the dynamic correlation between cross-modal features.

(2) We employ a heterogeneous graph generation module, which enables the graph network to aggregate both global visual and local semantic node features, thereby improving the recognition of flexible objects.

(3) We introduce the larger-scale Flexible Dataset (FSCW, \href{https://drive.google.com/file/d/1URk_mO4ucL1SLnNQ1YMzZvIxI_wtpSKK/view?usp=sharing}{access link}) to thoroughly evaluate the performance. We validate the effectiveness of the proposed method with extensive experiments on FDA, FSCW, CIFAR-100 and ImageNet-Hard datasets, achieving comparable performance.

% \begin{figure*}[ht]
%   \centering
%   \includegraphics[width=0.90\linewidth]{network.pdf}
%   \caption{The overall view of our proposed method. We employ adaptive scanning to improve the state-space model’s ability to capture contextual relationships between semantic nodes. Heterogenous graph relation learning is utilized to align semantic and visual nodes, strengthening the dynamic correlation between cross-modal features. Additionally, heterogeneous graph generation allows the graph neural network to effectively fuse global visual and local semantic node features.}
% \end{figure*}

\section{Related Work}
\subsection{Flexible Object Recognition}
% Although several studies \cite{jadon2019firenet, cheng2023convolution} have attempted to recognize flexible objects, they typically focus on one or two specific types rather than addressing the need for distinction and identification of multiple classes. For example, \cite{jadon2019firenet} proposed a CNN-based lightweight fire recognition algorithm, but it was designed solely for fire versus non-fire classification. CNNs \cite{jadon2019firenet} treat images as grid-like structures, where pixels or patches occupy fixed spatial positions, making these models less adaptable to the diverse shapes and sizes of flexible objects. In contrast, \cite{cheng2023convolution} proposed a vision transformer model for smoke recognition. By treating smoke image as sequences, this model captures dependencies among patches and improves recognition performance. Despite this improvement, the transformer method overlooks the inherent structural relationships between the patches. \cite{yang2025flexible} proposed a graph-vision method to better capture the relationships between nodes for flexible object recognition, but it only considers global visual node information, neglecting local semantic information. In contrast, our method proposes heterogeneous graph generation, which simultaneously captures both global visual and local semantic features.

Existing studies on flexible object recognition \cite{jadon2019firenet, cheng2023convolution} typically focus on one or two specific types rather than addressing the need for distinction and identification of multiple classes. For instance, \cite{jadon2019firenet} introduced a CNN-based lightweight fire recognition algorithm, limited to fire versus non-fire classification. While CNNs \cite{jadon2019firenet} treat images as grid-like structures, their fixed spatial positions hinder adaptability to diverse shapes and sizes of flexible objects. Vision transformers \cite{cheng2023convolution} improved smoke recognition by modeling patches as sequences to capture dependencies but overlooked inherent structural relationships. Similarly, graph-vision methods \cite{yang2025flexible} focus on global visual nodes while neglecting local semantic features. In contrast, our method proposes heterogeneous graph generation, which captures both global visual and local semantic features simultaneously.

\subsection{Graph Vision Model}
% \cite{han2022vision} proposed the Vision Graph Neural Network (ViG). This network treats each image patch as an individual graph node and uses a k-NN approach to establish relationships between these patches. Although this method is innovative, ViG \cite{han2022vision} primarily captures patch similarity, potentially overlooking the latent manifold structure of the image. Furthermore, the construction of graph structures is crucial, and Li \cite{li2019deepgcns} emphasized the importance of studying different distance metrics to build graph structures. However, ViG \cite{han2022vision} adopts the Euclidean distance to measure node dependencies for graph construction. This method fail to adequately capture the complex geometric relationships between image patches \cite{li2019deepgcns}. \cite{yao2022ml} uses cosine similarity to calculate the similarity between visual nodes and label nodes for cross-modal graph construction. Nonetheless, similar to the Euclidean distance in ViG, this method fails to align semantic and visual information and learn the dynamic correlation between cross-modal features, effectively. However, our method introduces heterogeneous graph relation learning, which better aligns semantic and visual nodes, enhancing the dynamic correlation between cross-modal features.

Vision Graph Neural Networks (ViG) \cite{han2022vision} treats image patches as graph nodes and uses k-NN to establish relationships between patches. While innovative, ViG relies on Euclidean distance, which overlooks the latent manifold structure and complex geometric relationships \cite{li2019deepgcns}. Similarly, \cite{yao2022ml} employs cosine similarity for cross-modal graph construction, but like the Euclidean distance in ViG, it fails to align semantic and visual information effectively or capture dynamic cross-modal correlations. In contrast, our method employs semantic-enhanced heterogeneous graph learning, which better aligns semantic and visual nodes, thereby enhancing the dynamic correlation between cross-modal features.

\subsection{State Space Model}
% The State-Space Model (SSM) \cite{huang2024localmamba}, known for its outstanding performance \cite{zhou2024mambainmambacentralizedmambacrossscantokenized} and excellent local feature extraction capabilities \cite{huang2024localmamba}, has garnered significant attention. Recent studies have introduced predefined scanning strategies \cite{zhou2024mambainmambacentralizedmambacrossscantokenized, huang2024localmamba} that map 2D images into 1D sequences, thus incorporating SSM into visual tasks. However, these fixed scanning strategies are not suitable for flexible objects with varying shapes and sizes. For example, Continuous Scan \cite{zhou2024mambainmambacentralizedmambacrossscantokenized} suffer from discontinuities in the neighbor node space, while Local Scan \cite{huang2024localmamba} suffers from semantic irrelevance. To this end, we propose an adaptive scanning method that allows the state space model to better capture the context between semantic nodes, thereby extracting more discriminative semantic features to match the varying shapes and sizes of flexible objects.

State-Space Models (SSMs) \cite{huang2024localmamba} are known for their local feature extraction capabilities but rely on predefined scanning strategies \cite{liu2024vmambavisualstatespace, huang2024localmamba} that are unsuitable for flexible objects. Strategies like Raster Scan \cite{liu2024vmambavisualstatespace} face discontinuities, while Local Scan \cite{huang2024localmamba} suffers from semantic irrelevance. To overcome these issues, we propose an adaptive scanning module that captures semantic context more effectively, enabling the extraction of discriminative features for objects with varying shapes and sizes.

\section{Method}
\subsection{Preliminaries}
\textbf{Graph Structure.} Unlike traditional GCN methods \cite{kipf2016semi} that treat the entire image or text as a single node, \cite{han2022vision} divide an image into \begin{math} {N} \end{math} patches and utilize graph embedding to associate them. And then, nodes vector \begin{math} V=[v_1,v_2,\ldots,v_N] \end{math} are obtained. For each node \begin{math} {v}_{i} \end{math} within the image, the Euclidean Distance is used to identify the \begin{math} {K} \end{math} nearest neighbors, collectively denoted as \begin{math} \mathcal{N}(v_i) \end{math}. These neighbor-nodes features are defined as \begin{math} e_{v}=[e_{v1},e_{v2,},\ldots,e_{v_N,}] \end{math}. For each neighboring node \begin{math} {v}_{j} \end{math}, an edge \begin{math} {e}_{ij} \end{math} is established to connect it to \begin{math} {v}_{i} \end{math}. By traversing all nodes, the set of edges \begin{math} {E} \end{math} can be obtained. Consequently, the graph can be constructed as \begin{math} {G=(V,E)} \end{math}.

\textbf{State Space Model.} The State Space Model (SSM) is a foundational model widely used to describe the dynamic behavior of a system, with applications in areas such as time series analysis and control systems. In SSM, the continuous evolution of the system is modeled through a set of ordinary differential equations (ODEs), which map an input signal to a latent space and then decode it into an output sequence. This process can be defined as:
\begin{equation}
\begin{cases}
h^{\prime}(t)=Ah(t)+Bx(t) \\
 % \\
y(t)=Ch(t) &
\end{cases},
\end{equation}
where, \begin{math} h(t) \end{math} represents the hidden state, while \begin{math} x(t) \end{math} and \begin{math} y(t) \end{math} denote the input and output, respectively. \begin{math} h^{\prime}(t) \end{math} represents the time derivative of \begin{math} h(t) \end{math}. Additionally, \begin{math} A \end{math} is the state transition matrix, and \begin{math} B \end{math} and \begin{math} C \end{math} are projection matrices. To discretize the continuous state space model, Mamba \cite{gu2023mamba} adopts a Zero-Order Hold (ZOH) technique to discretize the ordinary differential equations, as defined follows:
\begin{equation}
\begin{cases}
\overline{A}=\exp(\Delta A) \\
 % \\
\overline{B}=\left(\Delta A\right)^{-1}(\exp(\Delta A)-I)\cdot\Delta B & 
\end{cases},
\end{equation}
where, \begin{math} \Delta \end{math} is the time scale parameter, and \begin{math} \overline{A} \end{math} and \begin{math} \overline{B} \end{math} are the discretized forms of \begin{math} A \end{math} and \begin{math} B \end{math}, respectively. The discretized state space model can be expressed as:
\begin{equation}
\begin{cases}
h_t=\overline{A}h_{t-1}+\overline{B}x_t \\
 % \\
y_t=Ch_t & 
\end{cases}.
\end{equation}

\subsection{Adaptive Scanning Learning}
SSM, known for its excellent performance \cite{liu2024vmambavisualstatespace} and local feature extraction capabilities \cite{huang2024localmamba}, has attracted considerable attention. Building on this foundation, we propose an adaptive scanning method to improve feature extraction for flexible objects with varying shapes and sizes. For any two visual nodes \begin{math} v_i\in \mathbb{R}^{B\times D} \end{math} and \begin{math} v_j\in \mathbb{R}^{B\times D} \end{math}, we calculate their Euclidean distance and then take its reciprocal to evaluate the similarity between \begin{math} v_{i} \end{math} and \begin{math} v_{j} \end{math}:
\begin{equation}
s_{ij} = \frac{1}{d_{ij}} = \frac{1}{\sqrt{\sum_{k=1}^{D} \left(x_{ik} - x_{jk}\right)^2}},
\end{equation}
where, \begin{math} B \end{math} is batchsize and \begin{math} D \end{math} denotes dimension of nodes feature. The similarity \begin{math} s_{ij} \end{math} can be interpreted as the weight of the edge connecting two nodes. The set of weights is defined as \begin{math} W=[s_{11},s_{12},s_{1N},...,s_{NN}] \end{math}, enabling the construction of a weighted graph \begin{math} G=(V,E,W) \end{math}.

Furthermore, we apply the maximum spanning tree algorithm \cite{arogundade2011prim} to sort edges based on similarity weights from the graph \begin{math} G \end{math}, thereby constructing a maximum spanning tree \begin{math} T=(V,E_{T}) \end{math}. Here, \begin{math} E_{T}\subseteq E \end{math} represents the edge set of the maximum spanning tree, which determines a scanning path \begin{math} P=(v_{1}^{\prime},v_{2}^{\prime},...,v_{N}^{\prime})\mid v_{j}^{\prime}\in \mathbb{R}^{B\times D} \end{math}, where \begin{math} v_{1}^{\prime} \end{math},\begin{math} v_{2}^{\prime} \end{math},\begin{math} v_{N}^{\prime} \end{math} are the sorted nodes. Compared to scanning methods such as Raster Scan \cite{liu2024vmambavisualstatespace} and Local Scan \cite{huang2024localmamba}, our scanning method takes into account the correlations between nodes, incorporating richer contextual information to adapt to changes in the shape and size of flexible objects. By rewriting the state-space model in Eq.3, we obtain:
% \begin{equation}
% \begin{cases}
% h_j=\overline{A}^{\prime}h_{j-1}+\overline{B}^{\prime}{v_j}^{\prime} \\
%  % \\
% y_j=C^{\prime}h_j+{v_j}^{\prime} & &
% \end{cases},
% \end{equation}
\[
\begin{cases}
\begin{aligned}
  h_j &= \overline{A}'\,h_{j-1} + \overline{B}'\,v_j',\\
  y_j &= C'\,h_j + v_j'
\end{aligned}
\end{cases}
\]
where, \begin{math} \overline{A}^{\prime} \end{math}, \begin{math} \overline{B}^{\prime} \end{math} and \begin{math} C^{\prime} \end{math} represent the trainable parameters. After the SSM encoding, we obtain the semantic node vector \begin{math} V_s=[y_1,y_2,\ldots,y_N] \in \mathbb{R}^{B \times N \times D} \end{math}. Due to the adaptive scanning mechanism, the final semantic node vector \begin{math} V_s \end{math} incorporates locally correlated contextual information. 

\subsection{Semantic-Enhanced Heterogenous Graph Learning}
As illustrated in Fig.2. To address the heterogeneity between visual and semantic features, we design a flexible mapping module. This module dynamically generates the mapping matrices \begin{math} W_{v}\in \mathbb{R}^{B\times N\times D^{\prime}} \end{math} and \begin{math} W_{s}\in \mathbb{R}^{B\times N\times D^{\prime}} \end{math} from the input visual node vector \begin{math} V_v\in\mathbb{R}^{B\times N\times D} \end{math} and semantic node vector \begin{math} V_s\in \mathbb{R}^{B\times N\times D} \end{math}, enabling the dynamic alignment of feature spaces:
\begin{equation}
\begin{cases}
W_v=Conv(V_v)\cdot W_G \\ 
% \\
W_s=Conv(V_s)\cdot W_G
\end{cases},
\end{equation}
where, the graph embedding described in Section III.A \cite{han2022vision} is utilized, \begin{math} \mathrm{Conv}(\cdot) \end{math} denotes \begin{math} 1 \times 1 \end{math} convolution, which facilitates the mapping of heterogeneous features into a unified space. \begin{math} W_G\in \mathbb{R}^{D\times D^{\prime}} \end{math} is a globally shared learning parameter that ensures cross-modal consistency. \begin{math} B \end{math} represents the batch size, \begin{math} N \end{math} denotes the number of nodes, \begin{math} D \end{math} and \begin{math} D^{\prime} \end{math} correspond to the feature dimensions. This flexible mapping mechanism helps bridge the semantic gap and dynamically adjusts the importance of each node's features during fusion.

Additionally, since visual and semantic features may have different semantic meanings in various contexts, a dynamic context encoder is introduced to capture contextual information, thereby enhancing the model's representation ability across heterogeneous modalities:
\begin{equation}
\begin{cases}
C_v=MLP_v(V_v) \\ 
C_s=MLP_s(V_s)
\end{cases},
\end{equation}

\begin{table}[ht]
\centering
\caption{Comparison results of ours with current SOTA models on FDA.}
\resizebox{0.5\textwidth}{!}{%
\setlength{\tabcolsep}{6pt} % Adjust the column spacing as needed
\begin{tabular}{l|lccc}
% \toprule
\hline
\textbf{Category} & \textbf{Method} & \textbf{Parameters} & \textbf{Computation} & \textbf{Accuracy} \\
&  &  \textbf{(MB)} & \textbf{(GFLOPs)} & \textbf{(\%)} \\
\hline
% \midrule
Transformer & ViT-B/16 \cite{dosovitskiy2020image} & 86.80 & 17.60 & 76.38 \\
 & T2T-ViT-14 \cite{yuan2021tokens} & 21.50 & 4.80 & 75.46 \\
 & DaViT-Small \cite{ding2022davit} & 49.70 & 8.80 & 77.19 \\
 & Twins-SVT-S \cite{chu2021twins} & 24.00 & 2.90 & 30.32 \\
 & MixMAE \cite{liu2023mixmae} & 29.36 & 5.34 & 77.85 \\
% \hline
CNN 
 % & Resnet50 \cite{he2016deep} & 25.60 & 4.10 & 70.37 \\
 % & Resnet101 \cite{he2016deep} & 45.00 & 7.90 & 72.18 \\
 & Regnet \cite{radosavovic2020designing} & 4.78 & 0.41 & 74.65 \\
 & Densenet \cite{huang2017densely} & 18.00 & 4.37 & 71.29 \\
 & FireNet \cite{jadon2019firenet} & 0.62 & 0.01 & 54.05 \\
 & VAN-B0 \cite{guo2023visual} & 4.10 & 0.90 & 77.19 \\
% \hline
MLP & Mlp-mixer-base \cite{tolstikhin2021mlpmixer} & 59.00 & 12.70 & 68.76 \\
 & Mlp-mixer-larger \cite{tolstikhin2021mlpmixer} & 207.00 & 44.80 & 68.17 \\
\hline
Graph & Vig-s \cite{han2022vision} & 22.70 & 4.50 & 74.68 \\
\rowcolor{gray!20} % 设置这一行的背景颜色为浅灰色
 & \textbf{Ours} & \textbf{26.93} & \textbf{5.64} & \textbf{80.50} \\
\hline
\end{tabular}%
}
\end{table}

where, \begin{math} MLP_v \end{math} and \begin{math} MLP_s \end{math} represent the encoders that extract the visual context \begin{math} C_v\in \mathbb{R}^{B\times N\times D^{\prime}} \end{math} and semantic context \begin{math} C_s\in \mathbb{R}^{B\times N\times D^{\prime}} \end{math}, respectively. The contextual information is then introduced into the correlation learning to further enhance the cross-modal dynamic correlation between visual and semantic features:
\begin{equation}
\alpha_{ij}=soft\max(\frac{(W_v V_v^i+C_v)(W_sV_s^j+C_s)^\mathrm{T}}{\sqrt{d_a}}),
\end{equation}

where, \begin{math} \sqrt{d_a} \end{math} serves as the scaling factor, which prevents the numerator from becoming excessively large, thereby maintaining training stability.

Moreover, the cross-modal correlation matrix \begin{math} \alpha_{ij} \end{math} is passed through the nonlinear activation function LeakyReLU to enhance feature diversity. This allows for the calculation of graph heterogeneous correlation:
\begin{equation}
H_{ij}=\frac{\exp(\mathrm{LeakyReLU}(\alpha_{ij}))}{\sum_{k\in N}\exp(\mathrm{LeakyReLU}(\alpha_{ik}))}.
\end{equation}

\subsection{Heterogenous Graph Generation}
We propose a heterogeneous graph construction method that connects visual and semantic nodes. The visual nodes capture global information, while the semantic nodes provide local information. The fuse of both global and local information through heterogeneous graph learning enhances the recognition of flexible objects. Specifically, we treat the visual node \begin{math} V_{vi}\in V_{v}\mid V_{v}=\{v_{v1},v_{v2},...,v_{v N}\}\in \mathbb{R}^{B\times N\times D} \end{math} as the central node and use the K-Nearest Neighbors (KNN) algorithm to identify \begin{math} k \end{math} neighboring nodes in the semantic node vector \begin{math} V_s=\{v_{s1},v_{s2},...,v_{sN}\}\in \mathbb{R}^{B\times N\times D} \end{math}. The Euclidean distance, following the approach in \cite{han2022vision}, is employed as the node-to-node distance metric. From this, we compute the adjacency matrix:
\begin{equation}
Adj=Top\text{-}K(eudist,k),
\end{equation}
where, \begin{math} eudist \end{math} denotes the Euclidean distance between nodes. By applying the heterogeneous correlation \begin{math} H_{ij} \end{math} from Eq.9, we obtain a trainable distance metric \begin{math} H_{ij}\cdot(-eudist) \end{math}, which fully accounts for the dynamic correlation between visual and semantic nodes. As a result, this yields an adaptive adjacency matrix: 
\begin{equation}
Adj_H=Top\text{-}K(H_{ij}\cdot(-eudist),k).
\end{equation}
Finally, the heterogeneous graph is constructed as:
\begin{equation}
G_H=\begin{pmatrix}
V_v,V_s,E,Adj_H
\end{pmatrix}.
\end{equation}

\begin{table}[ht]
\centering
\caption{Comparison results with current SOTA models on FSCW.}
\resizebox{0.45\textwidth}{!}{%
\setlength{\tabcolsep}{16pt} % Adjust the column spacing as needed
\begin{tabular}{l|lccc}
% \toprule
\hline
\textbf{Category} & \textbf{Method} &  \textbf{Accuracy} \\
 &  &   \textbf{(\%)} \\
\hline
% \midrule
Transformer & DaViT-Small \cite{ding2022davit} & 86.91 \\
 & Twins-SVT-S \cite{chu2021twins} & 86.52 \\
 % & DeiT-S \cite{touvron2021training} & 86.14 \\
 & MixMAE \cite{liu2023mixmae} & 84.50 \\
 & RevViT \cite{mangalam2022reversible} & 80.12 \\
 % & MViT \cite{fan2021multiscale} & 88.06 \\
% \hline
MLP & Mlp-mixer-base \cite{tolstikhin2021mlpmixer} & 83.55 \\
 & Mlp-mixer-larger \cite{tolstikhin2021mlpmixer} & 82.02 \\
% \hline
CNN %& Resnet50 \cite{he2016deep} & 80.03 \\
 % & Regnet \cite{radosavovic2020designing} & 86.77 \\
 % & FireNet \cite{jadon2019firenet} & 71.90 \\
 & VAN-B0 \cite{guo2023visual} & 87.03 \\
 % & HorNet \cite{rao2022hornet} & 87.36 \\
 & FEC-Small \cite{chen2024neural} & 88.49 \\
 & RepLKNet \cite{ding2022scaling} & 81.25 \\
\hline
% MLP & Mlp-mixer-base \cite{tolstikhin2021mlpmixer} & 59.00 & 12.70 & 83.55 \\
%  & Mlp-mixer-larger \cite{tolstikhin2021mlpmixer} & 207.00 & 44.80 & 82.02 \\
% \hline
Graph & Vig-s \cite{han2022vision} & 88.21 \\
 % & Vig-ti \cite{han2022vision} & 87.42 \\
\rowcolor{gray!20} % 设置这一行的背景颜色为浅灰色
 & \textbf{Ours} & \textbf{89.98} \\
% \rowcolor{gray!20} % 设置这一行的背景颜色为浅灰色
%  & \textbf{Ours-tiny} & \textbf{88.23} \\
\hline
\end{tabular}%
}
\end{table}

\subsection{Graph Reason Learning}
Building upon the constructed heterogeneous graph, we will explore the interactions between the central node and its neighboring nodes. This process involves aggregation and updating steps designed to enhance the learning of node features.
\begin{equation}
G^{\prime}=Update(Aggregate(G_H,W_{agg}),W_{update}).
\end{equation}
Specifically, let \begin{math} G^{\prime} \end{math} denotes the updated graph, \begin{math} {W}_{agg} \end{math} and \begin{math} {W}_{update} \end{math} are the parameters for aggregation and updating, respectively. The graph convolution comprises \begin{math} l \end{math} layers (with \begin{math} l \end{math} = 16 in this paper), and the outputs of the upper layers serve as inputs to the lower layers through stacking. Finally, to increase the diversity of features, we utilize a feed-forward neural network (FFN) to map the node features.

‌%I、II、III、IV、V

\section{Experiments}
\subsection{Implementation Details \& Datesets}
% In the experimental configuration of our method, we fine-tuned various hyperparameters to enhance performance. The initial learning rate was established at 1.25e-4 and a linear warm-up with the ratio of 1e-4 is used. our experiments utilized an NVIDIA 3090 GPU with a batch size of 16. 
% Our model completed 100 epochs of training using the AdamW optimizer. The learning rate followed a cosine schedule, known for promoting efficient convergence. 
% 异构邻居节点数设置为9，空洞率为4用于模型去感知到更广泛的空间关系。
% To mitigate overfitting, we applied a dropout rate of 0.1. The architecture of the model featured 12 adjacent nodes and adopted a dynamic dilation rate that escalated with each layer's depth, increasing by 1 every 4 layers. This method of adaptive dilation allowed the model to perceive a wider range of spatial relationships.

Our model is trained using the Adam optimizer with an initial learning rate of 
1.25e-4, and a linear warm-up with a ratio of 1e-4. Training is conducted for 100 epochs with a batch size of 64 on an NVIDIA 3090 GPU. To mitigate overfitting, a dropout rate of 0.1 is applied. The number of heterogeneous neighbor nodes is set to 9, and a dilation rate of 4 is adopted to enable the model to perceive broader spatial relationships.

%https://anonymous.4open.science/r/FSCW-2EDE/

We conducted extensive experiments on the flexible datasets FDA \cite{yang2025flexible} and FSCW, as well as the general challenge datasets ImageNet-Hard \cite{taesiri2024imagenet} and CIFAR-100 \cite{krizhevsky2009learning}. The FDA dataset contains 2,474 images spanning 9 classes, including clouds, smoke, fog, fire, light, water, etc. Our proposed FSCW dataset includes 5,267 images across 6 classes, such as clouds, smoke-like clouds, smoke, fire, water, and sea, curated from CCSN \cite{zhang2018cloudnet}, Firenet \cite{jadon2019firenet}, CDD \cite{Niloy_2021}, and Kaggle. Recognizing flexible objects poses challenges due to their varying shapes, sizes, translucent attributes, and subtle inter-class differences. CIFAR-100 \cite{krizhevsky2009learning} consists of 60,000 images of size 32×32 distributed across 100 classes. ImageNet-Hard \cite{taesiri2024imagenet} is a curated subset of ImageNet, containing 8,822 images across 482 categories, selected for their complexity and difficulty.

% 我们在柔性数据集FDA和FSCW，以及通用挑战数据集ImageNet-Hard和CIFAR-100上开展了广泛的实验。其中FDA包含9个类（云、烟、雾、火、灯光、水面等）共2474张图片；FSCW包含6个类（云、类烟的云、烟、火、水面、海面）共5267张图片，从CCSN、Firenet、CDD \cite{Niloy_2021}、Kaggle。由于柔性目标形状和大小变化，透明特性，且类别之间差异较小这给识别带来了难度。CIFAR-100 包含100个类，大小为32x32，共60000张图片。ImageNet-Hard是从ImageNet精选的包含复杂性和挑战性的数据集，它有482个类别，共8822张图像。

%FDA 第一个柔性数据集 \cite{}
%FSCW 较大规模的柔性数据集
%IMAGENET-Hard & CIFAR-100 复杂挑战任务
%其他描述添加在补充材料
% We conduct extensive experiments on FDA \cite{yang2025flexible}, FSCW, CIFAR-100 \cite{krizhevsky2009learning} and ImageNet-Hard \cite{taesiri2024imagenet} datasets and detailed information about these datasets can be found in supplementary material.
% 我们在第一个柔性数据集和较大规模柔性数据集上展开了试验。
%FSCW 较大规模的柔性数据集

\begin{figure*}[ht]
  \centering
  \includegraphics[width=0.87\linewidth]{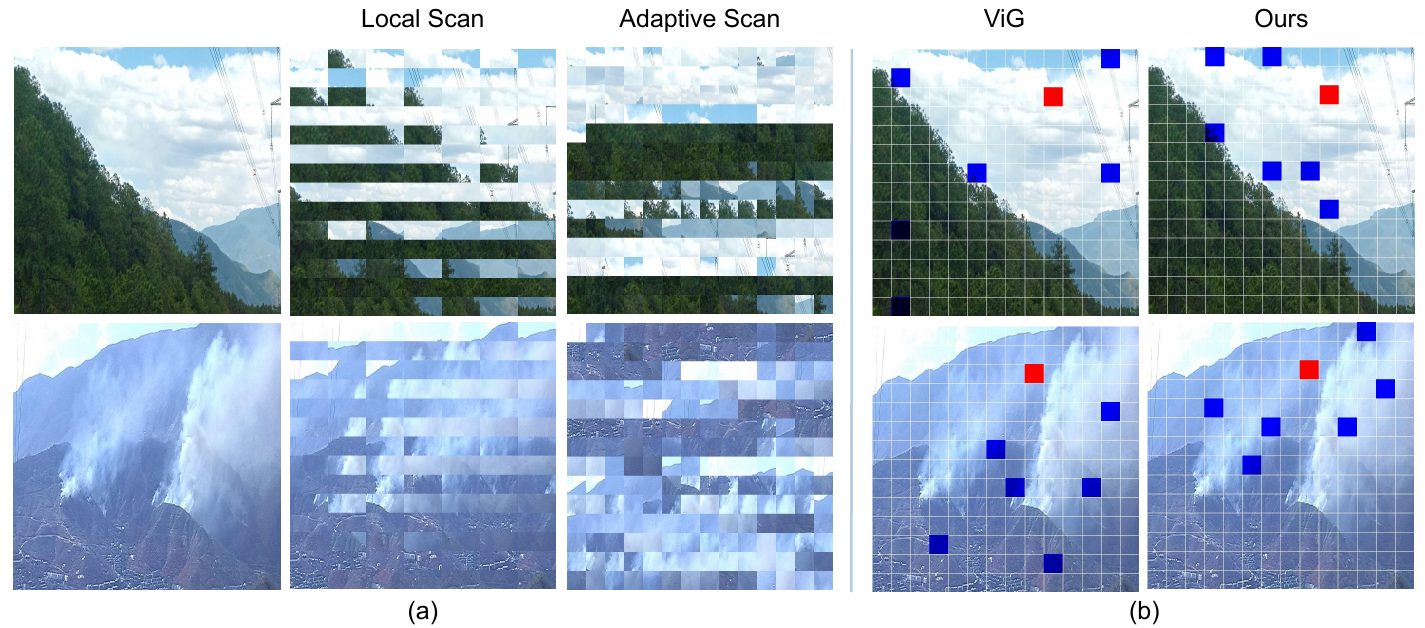}
  \caption{(a) Patch rearrangement results based on different scanning paths. The local scan method causes semantic irrelevance, while our adaptive scan method effectively captures discriminative semantic contexts. (b) Aggregation results of the central node (red, representing semantic information) and neighboring nodes (blue, representing visual information). Our method outperforms the ViG method in distinguishing flexible objects from the background.}
\end{figure*}

\subsection{Comparison with Other Methods}
% \textbf{FDA Dataset \cite{yang2025flexible}.} As shown in Table 1, Transformer-based models generally exhibit better accuracy. Notably, the MixMAE \cite{liu2023mixmae} model achieves an accuracy of 77.85\%, although it comes with relatively high computation and parameter requirements. In contrast, CNN-based models tend to perform worse, with most (e.g., ResNet101 \cite{he2016deep}, DenseNet \cite{huang2017densely}) exhibiting accuracy lower than that of Transformer and Graph-based models. FireNet \cite{jadon2019firenet}, while optimized for computational efficiency with fewer parameters, achieves a relatively low accuracy of only 54.05\%. Although VAN-B0 \cite{guo2023visual} strikes a balance between accuracy and computational cost, it still lags behind the proposed method by 3.4\%. MLP-based models, such as MLP-Mixer \cite{tolstikhin2021mlpmixer}, exhibit lower accuracy and are unsuitable for high-precision applications. In comparison, the proposed method achieves the highest accuracy of 80.50\%, while maintaining moderate computational and parameter demands, demonstrating the superior adaptability of Graph-based models for flexible object recognition.
% 我们的方法超过其他Transformer、CNN、MLP、Graph类的方法，在FDA和FSCW上分别取得最高的80.50%和89.98%的准确率，同时保持适当的计算量和参数量。这充分展示了我们方法
% 对形状大小变化，具有半透明特性的柔性目标识别的有效性及泛化性。
%I、II、III、IV、V
\textbf{Flexible Datasets.} As shown in Tables \text{I} and \text{II}, we conducted experiments on the flexible dataset FDA \cite{yang2025flexible} and our proposed larger-scale flexible dataset FSCW. Our method outperformed other Transformer, CNN, MLP, and Graph-based approaches, achieving the highest accuracy of 80.50\% on FDA and 89.98\% on FSCW, while maintaining moderate computational and parameter requirements. These results demonstrate the effectiveness and generalization ability of our method in recognizing flexible objects with varying shapes, sizes, and translucent attributes.

% \textbf{FSCW Dataset.} As shown in Table 2, within the Transformer category, DaViT-Small \cite{ding2022davit} and Twins-SVT-S \cite{chu2021twins} perform well, achieving accuracy above 86\%, while RevViT \cite{mangalam2022reversible} exhibits relatively lower performance with an accuracy of only 80.12\%. The MLP category overall shows slightly lower performance than the Transformer category, with Mlp-mixer-base and Mlp-mixer-larger \cite{tolstikhin2021mlpmixer} achieving accuracies below 84\%. In the CNN category, FEC-Small \cite{chen2024neural} achieves the highest accuracy at 88.49\%, surpassing Transformer methods, which suggests that CNNs are more effective at capturing local detail features, beneficial for flexible object recognition. In the Graph category, our method surpasses Vig-s \cite{han2022vision} with an accuracy of 89.98\%, the highest among all methods, demonstrating the strong adaptability of our method for flexible objects recognition tasks.

\begin{table}[ht]
\centering
\caption{Comparison results of ours with existing models on ImageNet-Hard and CIFAR-100 datasets.}
\Large
% \huge
\resizebox{0.42\textwidth}{!}{%
\setlength{\tabcolsep}{6pt} % Adjust the column spacing as needed
\begin{tabular}{l|lcc}
\hline
Category & Method & Accuracy (\%) & Accuracy (\%) \\
 & & ImageNet-Hard & CIFAR-100 \\
\hline
Transformer & Twins-SVT-S \cite{chu2021twins} & 62.74 & 71.67 \\
& T2T-ViT-14 \cite{yuan2021tokens} & 62.14 & 44.74 \\
& DeiT-S\cite{touvron2021training} & 62.50 & 71.14 \\
%\midrule
CNN & VAN-B0 \cite{guo2023visual} & 64.57 & 72.91 \\
    & Densenet \cite{huang2017densely} & 29.07 & 72.95 \\
%\midrule
MLP & Mlp-mixer-base \cite{tolstikhin2021mlpmixer} & 51.01 & 58.96 \\
 & Mlp-mixer-larger \cite{tolstikhin2021mlpmixer} & 47.71 & 56.76 \\
%\midrule
\rowcolor{gray!20} % 设置这一行的背景颜色为浅灰色
Graph & \textbf{Ours} & \textbf{65.39} & \textbf{83.30} \\
\hline
\end{tabular}%
}
\end{table}

%I、II、III、IV、V
\textbf{General Datasets.} In addition to the flexible datasets, we conducted additional comparative experiments with widely-used public datasets such as CIFAR100 \cite{krizhevsky2009learning} and ImageNet-Hard \cite{taesiri2024imagenet} to further confirm the generalizability of our method. As illustrated in Table \text{III}, our method achieved the highest accuracy on both datasets, with 83.30\% on CIFAR-100 and 65.39\% on ImageNet-Hard. These excellent performance demonstrates not only its ability to handle general datasets like CIFAR100 but also more complex and difficult datasets like ImageNet-Hard. The consistent improvement across other methods highlights the robustness and adaptability of our model to diverse tasks.

% \textbf{ImageNet-Hard \cite{taesiri2024imagenet} \& CIFAR-100 \cite{krizhevsky2009learning} Datasets.} As shown in Table 3, Transformer models (e.g., Twins-SVT-S \cite{chu2021twins}, T2T-ViT-14 \cite{yuan2021tokens}, and DeiT-s \cite{touvron2021training}) exhibit similar performance on the ImageNet-Hard dataset, achieving accuracies around 62\%. However, on the CIFAR-100 dataset, T2T-ViT-14 \cite{yuan2021tokens} achieves only 44.74\%, significantly underperforming compared to other methods. CNN models (e.g., VAN-B0 \cite{guo2023visual} and ResNet50 \cite{he2016deep}) demonstrate relatively stable performance across both the ImageNet-Hard and CIFAR-100 datasets, with VAN-B0 \cite{guo2023visual} showing a distinct advantage. MLP models perform poorly on both datasets, with accuracies below 60\%. Our method achieves an accuracy of 65.39\% on ImageNet-Hard and excels on CIFAR-100 with an accuracy of 83.30\%, significantly outperforming all other methods. This demonstrates that our method not only handles flexible objects recognition tasks well but also exhibits strong generalization ability on general challenging tasks.

\begin{figure}[ht]
  \centering
  \includegraphics[width=\linewidth]{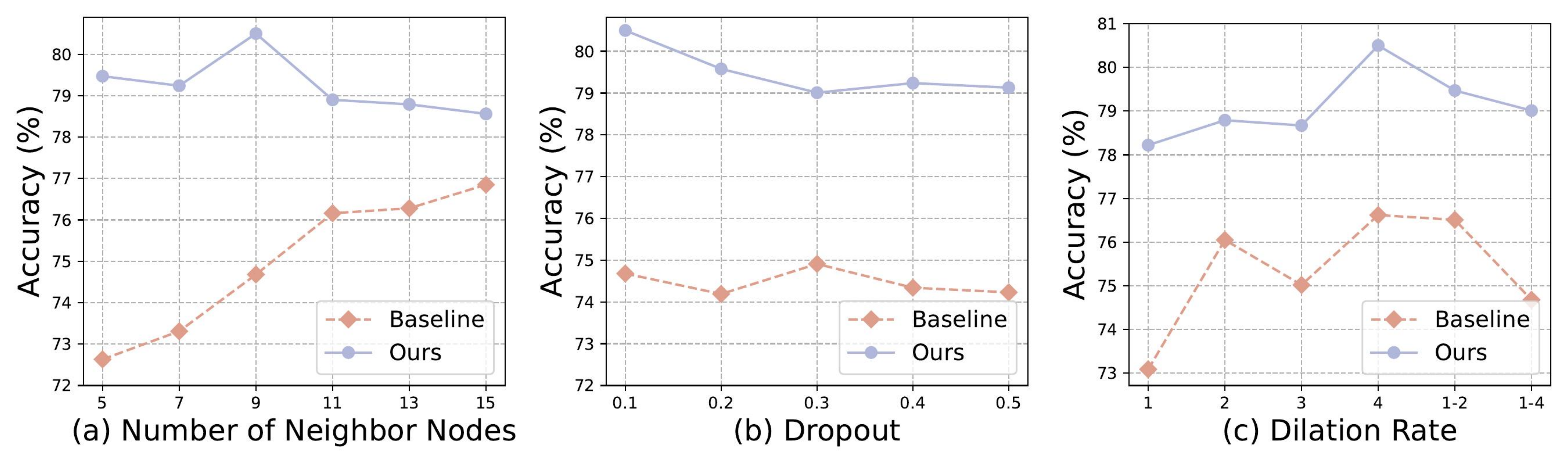}
  \caption{Influence of hyperparameters on performance.}
\end{figure}

\begin{table}[ht]
\centering
\caption{Ablation results of our method with various configurations.}
\resizebox{0.45\textwidth}{!}{%
\setlength{\tabcolsep}{6pt} % Adjust the column spacing as needed
\begin{tabular}{lcccc}
\hline
Baseline & Heterogenerous & Adaptive & Heterogenerous & Accuracy(\%) \\
         & Graph Generation & Scanning & Graph Learning & \\
\hline
\checkmark &  &  &  & 74.68 \\
\checkmark & \checkmark &  &  & 76.28(\textcolor{green}{\textcolor{green}{$\uparrow$}1.60})\\
\checkmark & \checkmark & \checkmark &  & 79.13(\textcolor{green}{\textcolor{green}{$\uparrow$}4.45}) \\
\checkmark &  & \checkmark & \checkmark & 79.01(\textcolor{green}{\textcolor{green}{$\uparrow$}4.33}) \\
\rowcolor{gray!20} % 设置这一行的背景颜色为浅灰色
\checkmark & \checkmark & \checkmark & \checkmark & 80.50(\textcolor{green}{\textcolor{green}{$\uparrow$}5.82}) \\
\hline
\end{tabular}
}
\end{table}

\subsection{Visualization}
Figure 3(a) visualizes the results of patch rearrangement of the original image based on different scanning paths. It can be observed that the local scan method leads to semantic irrelevance, where adjacent patches may represent different semantics. In contrast, our proposed adaptive scan method effectively captures discriminative semantic contexts. Figure 3(b) depicts the aggregation of the central node (red, representing semantic information) and neighboring nodes (blue, representing visual information). Compared to the ViG method, which only uses visual information, our method demonstrates superior capability in distinguishing flexible objects from their background.

\subsection{Influence of Hyperparameters}
As illustrated in Figure 4(a), as the number of neighbor nodes increases, our method consistently outperforms the baseline. Specifically when the number of neighbor nodes \begin{math} k=5 \end{math},  the baseline achieves only 72.63\% accuracy, while our method achieves 79.47\%. Furthermore, the accuracy of our method exhibits more stable fluctuations, demonstrating its insensitivity to variations in the number of neighboring nodes. In Figure 4(b), as the dropout rate varies, the baseline shows minor fluctuations in accuracy but remains at a relatively low level, while our method consistently outperforms it, showcasing the robustness of our approach. In Figure 4(c), our method surpasses the baseline across all dilation rates, with a notable improvement of 5.48\% at a dilation rate of 4.

\subsection{Ablation Study}
%I、II、III、IV、V
Building on the baseline, we achieved a 1.60\% improvement by incorporating heterogeneous graph generation alone, indicating that semantic node features provide richer local information. When both heterogeneous graph generation and adaptive scanning were integrated, a 4.45\% improvement over the baseline was observed in Table \text{IV}. This demonstrates that adaptive scanning enables the state-space model to better capture contextual relationships between semantic nodes, thereby facilitating the extraction of more discriminative semantic features. Finally, by combining all modules, we achieved a 5.82\% improvement over the baseline, highlighting that semantic-enhanced heterogeneous graph learning better aligns semantic and visual nodes, thereby enhancing the dynamic correlation between cross-modal features.

\section{Conclusion}
% This paper presents a novel approach to flexible object recognition, addressing challenges such as diverse shapes and sizes, ambiguous boundaries and subtle inter-class difference. We propose two key innovations: (1) a semantic-enhanced heterogeneous graph learning method that employs an adaptive scanning module, enabling the extraction of discriminative semantic context for matching flexible objects with varying shapes and sizes, while aligning semantic and visual nodes to enhance cross-modal feature correlation; (2) a heterogeneous graph generation method that aggregates global visual and local semantic node features to improve flexible objects recognition. Additionally, we employ the Flexible Object Recognition Dataset, FSCW, for evaluating our method. Extensive experiments on multiple datasets demonstrate that our approach outperforms existing methods, highlighting its effectiveness and adaptability for flexible object recognition.

This paper presents a novel approach to flexible object recognition, addressing challenges such as diverse shapes and sizes, ambiguous boundaries and subtle inter-class difference. We propose a semantic-enhanced heterogeneous graph learning method. Specifically, we employ an adaptive scanning module to extract discriminative semantic context, enabling the alignment of semantic and visual nodes to enhance cross-modal feature correlation. Moreover, we design a heterogeneous graph generation module to aggregate global visual and local semantic node features, further improving recognition accuracy. Additionally, we employ the Flexible Object Recognition Dataset, FSCW, for evaluating our method. Extensive experiments on multiple datasets demonstrate that our approach outperforms existing methods, highlighting its effectiveness and adaptability for flexible object recognition.

\bibliographystyle{IEEEbib}
\bibliography{main}

\begin{thebibliography}{10}

\bibitem{huang2017densely}
Gao Huang, Zhuang Liu, et~al.,
\newblock ``Densely connected convolutional networks,''
\newblock in {\em CVPR}, 2017, pp. 4700--4708.

\bibitem{radosavovic2020designing}
Ilija Radosavovic, Raj~Prateek Kosaraju, Ross Girshick, et~al.,
\newblock ``Designing network design spaces,''
\newblock in {\em CVPR}, 2020, pp. 10428--10436.

\bibitem{yuan2021tokens}
Li~Yuan et~al.,
\newblock ``Tokens-to-token vit: Training vision transformers from scratch on imagenet,''
\newblock in {\em Proceedings of the IEEE/CVF International Conference on Computer Vision}, 2021, pp. 558--567.

\bibitem{dosovitskiy2020image}
Alexey Dosovitskiy et~al.,
\newblock ``An image is worth 16x16 words: Transformers for image recognition at scale,''
\newblock {\em arXiv preprint arXiv:2010.11929}, 2020.

\bibitem{touvron2021training}
Hugo Touvron et~al.,
\newblock ``Training data-efficient image transformers \& distillation through attention,''
\newblock in {\em ICML}. 2021, pp. 10347--10357, PMLR.

\bibitem{ding2022davit}
Mingyu Ding, Bin Xiao, et~al.,
\newblock ``Davit: Dual attention vision transformers,''
\newblock in {\em European conference on computer vision}. Springer, 2022, pp. 74--92.

\bibitem{chu2021twins}
Xiangxiang Chu, Zhi Tian, et~al.,
\newblock ``Twins: Revisiting the design of spatial attention in vision transformers,''
\newblock {\em Advances in neural information processing systems}, vol. 34, pp. 9355--9366, 2021.

\bibitem{tolstikhin2021mlpmixer}
Ilya~O. Tolstikhin et~al.,
\newblock ``Mlp-mixer: An all-mlp architecture for vision,''
\newblock in {\em NeurIPS}, 2021.

\bibitem{jadon2019firenet}
Arpit Jadon, Mohd Omama, et~al.,
\newblock ``Firenet: a specialized lightweight fire \& smoke detection model for real-time iot applications,''
\newblock {\em arXiv preprint arXiv:1905.11922}, 2019.

\bibitem{zhang2018cloudnet}
Jiancheng Zhang, Ping Liu, et~al.,
\newblock ``Cloudnet: Ground-based cloud classification with deep convolutional neural network,''
\newblock {\em Geophysical Research Letters}, vol. 45, no. 16, pp. 8665--8672, 2018.

\bibitem{yang2025flexible}
Kunshan Yang, Lin Zuo, et~al.,
\newblock ``Flexible vig: Learning the self-saliency for flexible object recognition,''
\newblock {\em TCSVT}, 2025.

\bibitem{han2022vision}
Kai Han, Yunhe Wang, et~al.,
\newblock ``Vision gnn: An image is worth graph of nodes,''
\newblock {\em Advances in neural information processing systems}, vol. 35, pp. 8291--8303, 2022.

\bibitem{yao2022ml}
Ruijie Yao, Sheng Jin, et~al.,
\newblock ``Ml-vig: Multi-label image recognition with vision graph convolutional network,''
\newblock {\em Openreview}, 2022.

\bibitem{li2024deep}
Ziming Li, Bin Chen, et~al.,
\newblock ``Deep learning for urban land use category classification: A review and experimental assessment,''
\newblock {\em Remote Sensing of Environment}, vol. 311, pp. 114290, 2024.

\bibitem{huang2024localmamba}
Tao Huang, Xiaohuan Pei, et~al.,
\newblock ``Localmamba: Visual state space model with windowed selective scan,''
\newblock {\em arXiv preprint arXiv:2403.09338}, 2024.

\bibitem{liu2024vmambavisualstatespace}
Yue Liu, Yunjie Tian, et~al.,
\newblock ``Vmamba: Visual state space model,''
\newblock {\em arXiv preprint arXiv:2401.10166}, 2024.

\bibitem{cheng2023convolution}
Guangtao Cheng, Yancong Zhou, et~al.,
\newblock ``Convolution-enhanced vision transformer network for smoke recognition,''
\newblock {\em Fire Technology}, vol. 59, no. 2, pp. 925--948, 2023.

\bibitem{li2019deepgcns}
Guohao Li, Matthias Muller, et~al.,
\newblock ``Deepgcns: Can gcns go as deep as cnns?,''
\newblock in {\em Proceedings of the IEEE/CVF international conference on computer vision}, 2019, pp. 9267--9276.

\bibitem{kipf2016semi}
Thomas~N Kipf and Max Welling,
\newblock ``Semi-supervised classification with graph convolutional networks,''
\newblock {\em arXiv preprint arXiv:1609.02907}, 2016.

\bibitem{gu2023mamba}
Albert Gu and Tri Dao,
\newblock ``Mamba: Linear-time sequence modeling with selective state spaces,''
\newblock {\em arXiv preprint arXiv:2312.00752}, 2023.

\bibitem{arogundade2011prim}
OT~Arogundade, B~Sobowale, and AT~Akinwale,
\newblock ``Prim algorithm approach to improving local access network in rural areas,''
\newblock {\em International Journal of Computer Theory and Engineering}, vol. 3, no. 3, pp. 413, 2011.

\bibitem{liu2023mixmae}
Jihao Liu, Xin Huang, et~al.,
\newblock ``Mixmae: Mixed and masked autoencoder for efficient pretraining of hierarchical vision transformers,''
\newblock in {\em Proceedings of the IEEE/CVF Conference on Computer Vision and Pattern Recognition}, 2023, pp. 6252--6261.

\bibitem{guo2023visual}
Meng-Hao Guo, Cheng-Ze Lu, et~al.,
\newblock ``Visual attention network,''
\newblock {\em Computational Visual Media}, vol. 9, no. 4, pp. 733--752, 2023.

\bibitem{mangalam2022reversible}
Karttikeya Mangalam, Haoqi Fan, et~al.,
\newblock ``Reversible vision transformers,''
\newblock in {\em Proceedings of the IEEE/CVF Conference on Computer Vision and Pattern Recognition}, 2022, pp. 10830--10840.

\bibitem{chen2024neural}
Guikun Chen, Xia Li, et~al.,
\newblock ``Neural clustering based visual representation learning,''
\newblock in {\em Proceedings of the IEEE/CVF Conference on Computer Vision and Pattern Recognition}, 2024, pp. 5714--5725.

\bibitem{ding2022scaling}
Xiaohan Ding, Xiangyu Zhang, et~al.,
\newblock ``Scaling up your kernels to 31x31: Revisiting large kernel design in cnns,''
\newblock in {\em Proceedings of the IEEE/CVF conference on computer vision and pattern recognition}, 2022, pp. 11963--11975.

\bibitem{taesiri2024imagenet}
Mohammad~Reza Taesiri, Giang Nguyen, et~al.,
\newblock ``Imagenet-hard: The hardest images remaining from a study of the power of zoom and spatial biases in image classification,''
\newblock {\em Advances in Neural Information Processing Systems}, vol. 36, 2024.

\bibitem{krizhevsky2009learning}
Alex Krizhevsky, Geoffrey Hinton, et~al.,
\newblock ``Learning multiple layers of features from tiny images,''
\newblock {\em Computer Science University of Toronto}, 2009.

\bibitem{Niloy_2021}
Fahim~Faisal Niloy, Arif, et~al.,
\newblock ``A novel disaster image data-set and characteristics analysis using attention model,''
\newblock in {\em 2020 25th International Conference on Pattern Recognition (ICPR)}. Jan. 2021, p. 6116–6122, IEEE.

\end{thebibliography}

% \vspace{12pt}
% \color{red}
% IEEE conference templates contain guidance text for composing and formatting conference papers. Please ensure that all template text is removed from your conference paper prior to submission to the conference. Failure to remove the template text from your paper may result in your paper not being published.

\end{document}